# Requirements for Developing Robust Neural Networks

John S. Hyatt[a,b] and Michael S. Lee[a]

[a]Computational & Information Sciences Directorate, CCDC Army Research Laboratory, Aberdeen Proving Ground, MD 21005, USA
[b]Oak Ridge Associated Universities
john.s.hyatt11.ctr@mail.mil, michael.s.lee131.civ@mail.mil

**Abstract**

Validation accuracy is a necessary, but not sufficient, measure of a neural network classifier's quality. High validation accuracy during development does not guarantee that a model is free of serious flaws, such as vulnerability to adversarial attacks or a tendency to misclassify (with high confidence) data it was not trained on. The model may also be incomprehensible to a human or base its decisions on unreasonable criteria. These problems, which are not unique to classifiers, have been the focus of a substantial amount of recent research. However, they are not prioritized during model development, which almost always optimizes on validation accuracy to the exclusion of everything else. The product of this approach is likely to fail in unexpected ways outside of the training environment. We believe that, in addition to validation accuracy, the model development process must give added weight to other performance metrics such as explainability, resistance to adversarial attacks, and overconfidence on out-of-distribution data.

## 1. Introduction

Neural networks have proliferated in the past decade to the point that they are now a significant part of the digital ecosystem. They have been used to: build popular entertainment apps and web services (Anglade 2017, Paez 2019, Stark 2017), search engines (Clark 2015), and facial recognition software (Buolamwini 2018); allow actors to assume the appearance of politicians and celebrities (Zakharov 2019); operate autonomous vehicles (Bojarski et al. 2016); analyze satellite imagery and other long-range/remote sensing data (Zhu et al. 2017); and much more.

Paralleling the increasing presence of neural network-based products in everyday life is a growing public awareness of their potential shortcomings. Flawed machine learning models, many of them based on neural networks, comprise a significant fraction of all implementations and have been making headlines for years. Specific examples include racial (Angwin et al. 2016; Barr 2015; Buranyi 2017; Garvie et al. 2016; Hunt 2016) and gender (Day 2019) bias, broken trending algorithms (Thielman 2016), innocent people identified as criminals (Dodd 2018), and death (Marshall and Davies 2019). These are separate from incidents involving the *successful* application of machine learning for nefarious purposes, such as the Facebook/Cambridge Analytica scandal (Cadwalladr 2019).

The "standard" neural network development process optimizes for performance on a specific distribution of data, but does *not* optimize for generality, explainability, or robustness. These qualities have been the subject of a substantial amount of research, but despite the advances in this area the vast majority of work treats them as secondary considerations, if at all. We believe that many model failures are preventable and attribute them to (i) the machine learning community's tendency to compartmentalize robustness research, (ii) companies' failure to address robustness during development and marketing, and (iii) end users' low awareness of the potential risks.

Below, we describe what this means for the case of a neural network classifier trained on labeled data, the same case we examined quantitatively in (Hyatt and Lee 2019). None of the technical ideas are new: partial solutions to this problem already exist, and only need to be incorporated into the development process. We emphasize that the basic concepts also apply to other types of neural networks, such as unsupervised classifiers and regression models.

In this paper we take the position that in order to make models trustworthy, transparent, robust, and resilient, these qualities must be prioritized from the beginning of the development process. We specifically hope to make this case to non-technical policymakers and others considering the purchase, use, and regulation of neural networks, especially in the public sector and government. We also hope researchers will view this not only as an ethical concern but as an opportunity to strengthen the foundations of the field.

## 2. Example: Supervised Classifiers

Broadly speaking, most supervised classifiers are developed using some variation of the same basic procedure (Russell and Norvig 2009):

1. Collect a dataset containing many examples of each class to be identified, as well as their labels.
2. Split the data into training, validation, and test sets.
3. Choose a model architecture and training scheme (loss function, optimizer, etc.).
4. Iteratively update the model by feeding it the training data and adjusting its weights to maximize the fraction of the training set the model classifies correctly.
5. After each iteration, evaluate the classifier's accuracy on the validation set.
6. Repeat step 5 until the validation accuracy plateaus.
7. If the validation accuracy is unsatisfactory, repeat steps 3–6 with a different architecture and/or training scheme.
8. Evaluate the best model on the test set and report those results. Training is complete.

Throughout the training process, every major decision is driven by the value of one metric, in this case the validation accuracy. However, this sort of narrow evaluation, which ignores numerous sources of uncertainty in both the development process and the model's expected real-world performance, greatly oversimplifies the problem.

To begin with, the training data may be subject to *biases or constraints* that the real-world data is not. To take a high-profile example involving the public sector, consider commercial face-recognition software, which has been used with minimal restraint by police departments across the country despite substantial racial bias encoded into these products (Garvie et al. 2016). Face recognition models are frequently trained on an insufficiently diverse dataset with unbalanced race/gender demographics. Because the validation set comes from the same distribution of data as the training set, it masks this problem and prevents it from being addressed during the development process, leading to finished products that incorporate the racial and gender bias present in the data (Buolamwini and Gebru 2018; Phillips et al. 2011). Face recognition software developed in the West identifies white men with a high degree of accuracy, performing significantly worse with women and people of color, and worst of all with women of color. Similarly, Chinese software is very good at identifying Asian men, but not women or non-Asians. Depending on the specific metric, performance on dark-skinned women can be up to 10× worse than for light-skinned men (Buolamwini and Gebru 2018).

Truly *out-of-distribution data* presents a distinct but closely-related hazard (Hendrycks and Gimpel 2017; Lakshminarayanan et al. 2017; Lee et al. 2018; Li and Gal 2017). Continuing the previous example, how does face recognition software classify an image of something that is not a human face? Rather than failing to correctly identify classes underrepresented in the training data, such a classifier might make nonsensical predictions on data it was not trained to handle. By design, a classifier's predictions are restricted to the classes it was intended to identify (in this case, faces). However, it should not identify something that is not a human face with high confidence.

To identify problems like these in development, or pinpoint their causes when they occur in the wild, a human must be able to understand why the classifier makes the predictions it does. Similarly, a human using it must be able to trust that the classifier is making decisions in a rational way. In other words, a model should be *explainable,* at least to a specialist (and ideally to anyone using it).

Finally, even if the model is neither biased nor overconfident when presented with out-of-distribution data, it may be *vulnerable to an adversarial attack* (Yuan et al. 2017). The attacker identifies the shortest distance, in the model's internal representation of the input space, to the boundary separating the predicted class from another class, and perturbs the input data along that path. Seemingly negligible changes to the data, chosen in this way, can cause the model to misclassify an input. Adversarial attacks are notoriously difficult to prevent, even if the attacker does not have access to the details of the original model or the data used to train it (Papernot et al. 2017). Training the model on adversarially generated data does not provide a complete solution to the problem (Kurakin et al 2017), although it likely helps bring the training data closer to the real-world target distribution. This phenomenon can also overlap with out-of-distribution data, as it is possible to generate unrecognizable data that the model nevertheless identifies with high confidence (Nguyen et al. 2015).

There is a common theme tying these challenges together: *high validation accuracy does not imply that the model will overcome them.* Biased datasets are a recognized problem (although much work remains to be done in that area) but the roles of model design and optimization metrics are less explored. The standard model development process optimizes only validation accuracy, when it should additionally optimize for a family of *robustness* criteria: for example, confidence of predictions on out-of-distribution data, explainability, and resistance to adversarial attacks. Our research (Hyatt and Lee 2019) indicates that these criteria are related to one another and provide complementary windows into the quality of the model's internal representation of the data. For example, explainable models tend to be those that segment their input space with reasonable decision boundaries, and therefore are less susceptible to adversarial attacks. This overlap may also help offset the increased cost to optimize and test for robustness – it may be enough to optimize for the cheapest metric early on, then either use the others to refine the model or as independent tests of its robustness. For these reasons, because of variation within a given test (e.g., resistance to different types of adversarial attack), and because of the harms already occasioned by overemphasizing validation accuracy, we hesitate to promote a specific robustness metric. This is a complex problem, and without a complete theory of machine learning any metrics will be somewhat ad hoc.

We note that having a human in the loop is *not* a solution. In some cases this is because humans are not able to easily understand the type or volume of data; humans (even experts) also have a tendency to trust supposedly authoritative predictions; and having a human constantly monitor the network's output cancels out any benefits gained by automating the decision-making process.

## 3. Solutions

The ideas in the preceding section apply equally to neural networks other than supervised classifiers, and our suggested solution is the same in all cases. Independent of the values of the network weights, the network architecture determines both the amount of information that can be encoded in the model as well as the path that information takes through the network during training. Architecture is not the same as size: smaller networks with the correct architecture optimize much more effectively and are more robust than large, poorly-designed networks. However, a robust model must encode more (sometimes much more) information than one which merely performs well on the validation set. There are well-known trade-offs between accuracy and robustness for a given architecture, and also between accuracy or robustness and the architecture itself, in terms of cost (to train, to store, and to evaluate). Optimizing network design for robustness in addition to metrics such as validation accuracy will mean designing architectures that can efficiently encode sufficient information.

Because the network requires more information to make reasonable decisions in unusual circumstances (and therefore meet robustness criteria such as those outlined above), this will inevitably increase network size and training cost relative to a naïve network. This cost is in addition to any required to ensure that the training data distribution matches the real world. Designing a network that performs well on these metrics will cost much more than the current (insufficient) industry standard, resulting in "sticker shock" – especially since some robustness metrics are not easily captured by a single number. However, we believe that the benefits more than make up for the increased cost.

Finally, *designing* a network from scratch to solve an arbitrary problem is not possible at this point. Most model development is based off of heuristics, trial and error, and adapting existing solutions to similar problems. There is not a complete theoretical foundation for neural network model design.

The closest extant solution is the young field of automated machine learning (auto-ML), in which a meta-model is trained, itself in a non-explainable fashion, to build networks to solve particular problems. However, to our knowledge, these meta-models always define success solely in terms of the generated model's validation accuracy (Fusi et al. 2018; Jin et al. 2018; Zoph and Le 2017), meaning that the generated models will have all the shortcomings described above. In principle, there is no reason that the auto-ML optimization criteria cannot include other metrics.

We see three solutions to the problems described above, at least one of which should be applied to any neural network model, but especially those used in critical, public sector, and/or governmental applications:

- Incorporate robustness optimization, including existing measures, explicitly into the development process and verify that the finished model meets predetermined robustness standards. If it does not, the model architecture must be changed and the model re-trained, possibly with additional data. (In the training outline at the beginning of Section 2, this modifies step 7.)
- Expand current auto-ML meta-models to optimize for a number of robustness metrics in addition to validation accuracy.
- Develop our theoretical understanding of machine learning so that the requirements to solve an arbitrary problem (network architecture, training scheme, data attributes, etc.) can be determined from first principles at the beginning of the development process.

Of these, only the first can be implemented using currently available tools, though this will substantially increase development costs. As with the current process, heuristics learned over time will bring this cost down, but the process will always be more time-consuming than it is now.

The second could be achieved using nearly current technology with enough investment. Companies like Microsoft and Google (which already market auto-ML services) would need to be incentivized or required by law to do so.

The third is a long-term goal and will likely require years of research, and for the machine learning community to work closely with experts in fields like mathematics, physics, and information theory.

Regardless of the details, *there is a pressing need for comprehensive, industry-wide standards in this field.*

## 4. Conclusion

Current top-line numbers reported for neural networks (e.g., validation accuracy) overestimate those networks' strengths and downplay their weaknesses. Only a small fraction of neural network research papers verify the robustness of their models unless that is the explicit purpose of the paper; most machine learning competitions do not require this type of evaluation; and commercial machine learning products are not held to a comprehensive standard. Similarly, most end users of neural network products are unaware of the severity of these issues. It is critical to hold these products to high standards of accountability and transparency, particularly in the public sphere, and the only way to do that is to ensure that robustness is a primary consideration during model development.

# References

Anglade, T. 2017. How HBO's Silicon Valley Built "Not Hotdog" with Mobile TensorFlow, Keras & React Native. *Medium.*

Angwin, J.; Larson, J.; Mattu, S.; and Kirchner, L. 2016. Machine Bias. *ProPublica.*

Barr, A. 2015. Google Mistakenly Tags Black People as Gorillas, Showing Limits of Algorithms. *Wall Street Journal.*

Bojarski, M.; Testa, D. D.; Dworakowski, D.; Firner, B.; Flepp, B.; Goyal, P.; Jackel, L. D.; Monfort, M.; Muller, U.; Zhang, J.; Zhang, X.' Zhao, J.; and Zieba, K. 2016. End to End Learning for Self-Driving Cars. *Computing Research Repository* abs/1604.07316.

Buolamwini, J. and Gebru, T. 2018. Gender Shades: Intersectional Accuracy Disparities in Commercial Gender Classification. In Proceedings of the 1st Conference on Fairness, Accountability and Transparency, 77–91. New York, N.Y.: Proceedings of Machine Learning Research.

Buranyi, S. 2017. Rise of the Racist Robots – How AI is Learning All Our Worst Impulses. *The Guardian.*

Cadwalladr, C. 2019. "I Made Steve Bannon's Psychological Warfare Tool": Meet the Data War Whistleblower. *The Guardian.*

Clark, J. 2015. Google Turning its Lucrative Web Search Over to AI Machines. *Bloomberg.*

Day, M. 2019. How LinkedIn's Search Engine May Reflect a Gender Bias. *Seattle Times.*

Dodd, V. 2018. UK Police Use of Facial Recognition Technology a Failure, Says Report. *The Guardian.*

Fusi, N.; Sheth, R.; and Elibol, M. H. 2018. Probabilistic Matrix Factorization for Automated Machine Learning. In Advances in Neural Information Processing Systems 31, 3348–3357. Red Hook, N.Y.: Curran Associates, Inc.

Garvie, C.; Bedoya, A.; and Frankle, J. 2016. The Perpetual Line-Up: Unregulated Police Face Recognition in America. Georgetown Law, Center on Privacy & Technology.

Hendrycks, D. and Gimpel, K. 2017. A Baseline for Detecting Misclassified and Out-of-Distribution Examples in Neural Networks. Paper presented at 2017 International Conference on Learning Representations, Toulon, France, 24–26 April.

Hunt, E. 2016. Tay, Microsoft's AI Chatbot, Gets a Crash Course in Racism From Twitter. *The Guardian.*

Hyatt, J. S. and Lee, M. S. 2019. Beyond Validation Accuracy: Incorporating Out-of-Distribution Checks, Explainability, and Adversarial Attacks into Classifier Design. In Proceedings of SPIE 11006, Artificial Intelligence and Machine Learning for Multi-Domain Operations Applications, 110061L. Bellingham, Wash.: SPIE Press.

Jin, H.; Song, Q.; and Hu, X. 2018. Efficient Neural Architecture Search with Network Morphism. *Computing Research Repository* abs/1806.10282.

Kurakin, A.; Goodfellow, I. A.; and Bengio, S. 2017. Adversarial Machine Learning at Scale. Paper presented at 2017 International Conference on Learning Representations, Toulon, France, 24–26 April.

Lakshminarayanan, B.; Pritzel, A.; and Blundell, C. 2017. Simple and Scalable Predictive Uncertainty Estimation Using Deep Ensembles. In Advances in Neural Information Processing Systems 30, 6402–6413. Red Hook, N.Y.: Curran Associates, Inc.

Lee, K.; Lee, H.; Lee, K.; and Shin, J. 2018. Training Confidence-Calibrated Classifiers for Detecting Out-of-Distribution Samples. Paper presented at 2018 International Conference on Learning Representations, Vancouver, Canada, 30 April–3 May.

Li, Y. and Gal, Y. 2017. Dropout Inference in Bayesian Neural Networks with Alpha-Divergence. In Proceedings of the 34th International Conference on Machine Learning, 2052–2061. New York, N.Y.: Proceedings of Machine Learning Research.

Marshall, A. and Davies, A. 2018. Uber's Self-Driving Car Saw the Woman it Killed, Report Says. *Wired.*

Nguyen, A.; Yosinski, J.; and Clune, J. 2015. Deep Neural Networks are Easily Fooled: High Confidence Predictions for Unrecognizable Images. In IEEE Conference on Computer Vision and Pattern Recognition, 427–436. Silver Spring, Md.: IEEE Computer Society Press.

Paez, D. 2019. This Person Does Not Exist is the Best One-Off Website of 2019. *Inverse.*

Papernot, N.; McDaniel, P.; Goodfellow. I.; Jha, S.; Celik, Z. B.; and Swami, A. 2017. Practical Black-Box Attacks Against Machine Learning. In Proceedings of the 2017 ACM on Asia Conference on Computer and Communications Security, 506–519. New York, N.Y.: ACM.

Phillips, P. J.; Jiang F.; Narvekar, A.; Ayyad, J.; and O'Toole, A. J. 2011. An Other-Race Effect for Face Recognition Algorithms. *ACM Transactions on Applied Perception* 8(2): 14.

Russell, S. and Norvig, P. 2009. *Artificial Intelligence: A Modern Approach.* Upper Saddle River: Prentice Hall Press.

Stark, H. 2017. Introducing FaceApp: The Year of the Weird Selfies. *Forbes.*

Thielman, S. 2016. Facebook Fires Trending Team, and Algorithm Without Humans Goes Crazy. *The Guardian.*

Yuan, X.; He, P.; Zhu, Q.; Bhat, R. R.; and Li, X. 2017. Adversarial Examples: Attacks and Defenses for Deep Learning. *IEEE Transactions on Neural Networks and Learning Systems* 1–20.

Zakharov, E.; Shysheya, A.; Burkov, E.; and Lempitsky, V. 2019. Few-Shot Adversarial Learning of Realistic Neural Talking Head Models. *Computing Research Repository* abs/1905.08233.

Zhu, X. X.; Tuia, D.; Mou, L.; Xia, G.; Zhang, L.; Xu, F.; and Fraundorfer, F. 2017. Deep Learning in Remote Sensing: A Comprehensive Review and List of Resources. *IEEE Geoscience and Remote Sensing Magazine* 5(4): 8–36.

Zoph, B. and Le, Q. V. 2017. Neural Architecture Search with Reinforcement Learning. *Computing Research Repository* abs/1611.01578.